\documentclass[10pt,letterpaper]{paper}

\usepackage{lmodern} 
\usepackage{silence}                             
\usepackage[all]{hypcap}

\usepackage[authoryear, round]{natbib}

\usepackage[utf8]{inputenc} % allow utf-8 input
\usepackage[T1]{fontenc}    % use 8-bit T1 fonts
\usepackage{url}            % simple URL typesetting
\usepackage{booktabs}       % professional-quality tables
\usepackage{amsfonts}       % blackboard math symbols
\usepackage{nicefrac}       % compact symbols for 1/2, etc.
\usepackage{microtype}      % microtypography
\usepackage{multirow}
\usepackage[textsize=tiny]{todonotes}
\usepackage{algorithm}
\usepackage{algpseudocode}
\usepackage{amssymb}
\usepackage{cleveref}
\usepackage{dsfont}
\usepackage{nicefrac}
\usepackage{xcolor}
\usepackage{amsmath}
\usepackage{amssymb}
\usepackage{booktabs}
\usepackage{multirow}
\usepackage{makecell}
\usepackage{caption}
\usepackage{subcaption}
\usepackage{bbm}
\usepackage{wrapfig}
\usepackage{needspace}
\usepackage{float}
\usepackage{mathtools,xparse}
\usepackage{pifont}
\usepackage{enumitem}
\usepackage{array}
\usepackage{scalerel,xparse}
\usepackage{colortbl}
\usepackage{tablefootnote}
\usepackage{tocloft}
\usepackage{siunitx}

\usepackage{hyperref}
\newcommand{\eqautoref}[1]{\hyperref[#1]{Eq.~\ref*{#1}}}
\newcommand{\eqsautoref}[2]{\hyperref[#1]{Eqs.~\ref*{#1}} and~\hyperref[#2]{\ref*{#2}}}
\newcommand{\secautoref}[1]{\hyperref[#1]{Section~\ref*{#1}}}
\newcommand{\appautoref}[1]{\hyperref[#1]{Appendix~\ref*{#1}}}
\newcommand{\figautoref}[1]{\hyperref[#1]{Figure~\ref*{#1}}}
\newcommand{\figsautoref}[2]{\hyperref[#1]{Figures~\ref*{#1}} and~\hyperref[#2]{\ref*{#2}}}
\newcommand{\tabautoref}[1]{\hyperref[#1]{Table~\ref*{#1}}}
\newcommand{\algautoref}[1]{\hyperref[#1]{Algorithm~\ref*{#1}}}

\usepackage{dblfloatfix}
\usepackage{adjustbox}
\usepackage{setspace}
\usepackage[most,skins,theorems]{tcolorbox}
\usepackage{graphicx}

\newboolean{showsection}
\setboolean{showsection}{true}
\makeatletter
\@namedef{ver@everyshi.sty}{}
\makeatother

\definecolor{custom_green}{rgb}{0.0, 0.5, 0.0}
\definecolor{custom_red}{rgb}{1.0, 0.01, 0.24}
\definecolor{custom_blue}{HTML}{C9DAF7}
\definecolor{custom_purple}{HTML}{D9D1E9}
\definecolor{title_blue}{HTML}{204899}
\definecolor{cite_blue}{HTML}{044dc1}
\definecolor{cite_purple}{HTML}{7406a7}
\definecolor{microsoft_red}{HTML}{ec4e21}

\hypersetup{
    colorlinks = true,
    citecolor = {cite_blue},
    linkcolor = {cite_purple},
    urlcolor = {cite_purple},
}

\definecolor{blanchedalmond}{rgb}{1.0, 0.92, 0.8}
\definecolor{carmine}{rgb}{0.59, 0.0, 0.09}
\definecolor{lightblue}{rgb}{0.22,0.45,0.70}%

\renewcommand{\mathbf}{\boldsymbol}

\makeatletter
\def\Ddots{\mathinner{\mkern1mu\raise\p@
\vbox{\kern7\p@\hbox{.}}\mkern2mu
\raise4\p@\hbox{.}\mkern2mu\raise7\p@\hbox{.}\mkern1mu}}
\makeatother

\numberwithin{equation}{section}

\definecolor{amaranth}{rgb}{0.9, 0.17, 0.31}
\definecolor{antiquebrass}{rgb}{0.8, 0.58, 0.46}
\definecolor{antiquefuchsia}{rgb}{0.57, 0.36, 0.51}
\definecolor{chromeyellow}{rgb}{0.31, 0.47, 0.26}

\newcommand{\1}{\mathds 1}

\usepackage{amsmath,amsfonts,bm}

\def\eqref#1{equation~\ref{#1}}
\def\1{\bm{1}}

\DeclareMathAlphabet{\mathsfit}{\encodingdefault}{\sfdefault}{m}{sl}
\SetMathAlphabet{\mathsfit}{bold}{\encodingdefault}{\sfdefault}{bx}{n}

\makeatletter
\def\mathcolor#1#{\@mathcolor{#1}}
\def\@mathcolor#1#2#3{%
  \protect\leavevmode
  \begingroup
    \color#1{#2}#3%
  \endgroup
}
\makeatother

\Crefformat{equation}{#2Eq.\;(#1)#3}
\Crefformat{figure}{#2Figure #1#3}
\Crefformat{assumption}{#2Assumption #1#3}
\Crefname{assumption}{Assumption}{Assumptions}

\usepackage{crossreftools}
\makeatletter
\renewcommand\footnoterule{%
  \kern 15\p@
  \hrule \@width 2in \kern 2.6\p@
  \vspace{4pt}
}
\makeatother

\reportnumber{}

\title{
No Pain, More Gain: Iterative Merging for Effective Multi-Teacher On-Policy Distillation
}

\author[1,*]{Seonghyeon Kim}
\author[1,*]{Chaeyun Jang}
\author[2]{Noah Lee}
\author[2]{Boseop Kim}
\author[1]{Juho Lee}  

\makeatletter
\renewcommand{\AB@affilsepx}{,\protect\hspace{0.5em}\protect\Affilfont}
\makeatother

\affil[1]{KAIST AI}
\affil[2]{Kakao}

\correspondingauthor{%
Correspondence to:
{
\email{\{shkim0824, jcy9911, juholee\}@kaist.ac.kr}}.%
\newline
Code: {\url{https://github.com/shkim0824/immopd}}.%
}

\begin{abstract}

\textbf{Abstract:}
Multi-teacher on-policy distillation (MOPD) combines independently developed domain teachers into a single student by distilling their predictions on student-generated samples. We study a setting where teachers share a reference model but undergo different post-training procedures, and find that MOPD can struggle to recover some teacher capabilities. Because distillation occurs on student-generated prefixes, the student initialization can strongly affect subsequent recovery. However, initial benchmark performance is not a reliable predictor of a good MOPD initialization. For example, merge initialization can start below SFT warm-up yet finish higher after MOPD. We further find that effective merging depends on both the relative teacher contributions and the overall merge scale, with some strong configurations lying outside the simplex of convex parameter averaging. Thus, selecting a good merge initialization requires evaluating not only its immediate performance but also the learning it enables under MOPD, making one-shot coefficient search difficult. We propose \textbf{I}terative \textbf{M}erging for \textbf{MOPD} (\textbf{IM-MOPD}), which starts from a uniform merge and progressively adds task-vector increments for under-recovered domains during distillation. In a 5-domain setting, IM-MOPD achieves higher average normalized recovery than MOPD with either uniform merge initialization or SFT warm-up, showing that effective teacher contributions can be determined progressively during training.
\end{abstract}

\begin{document}

\maketitle

\section{Introduction}
\label{sec:intro}

\begin{figure*}[!h]
    \centering
    \includegraphics[width=\linewidth]{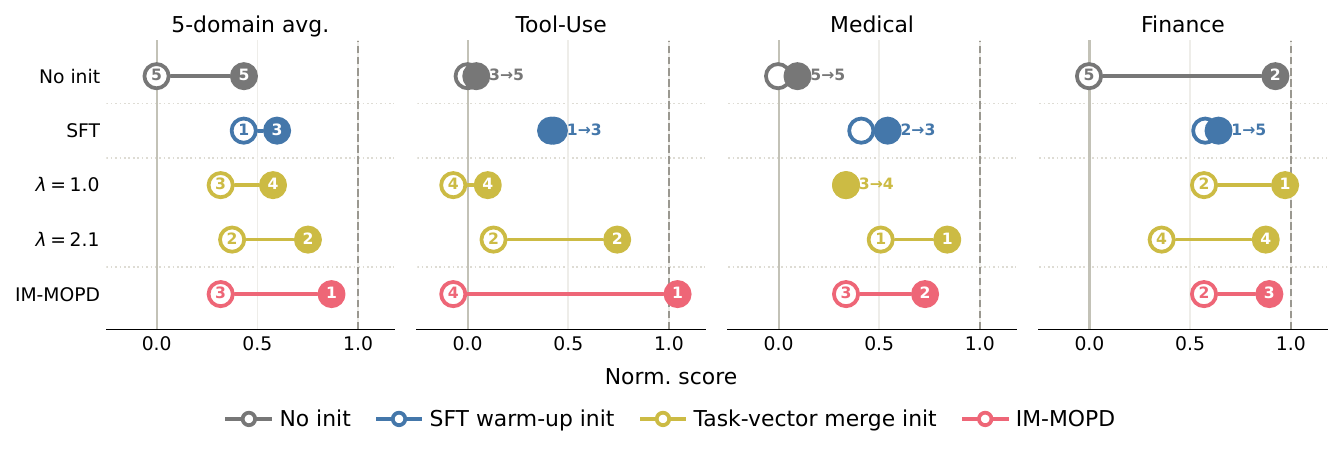}
    \caption{\textbf{Student initialization matters for MOPD, but stronger initial performance does not necessarily lead to better outcomes.} On Qwen3-4B in the 5-domain setting, MOPD initialized from the shared reference model struggles especially with the SFT-trained Tool Use and Medical teachers, while successfully recovering the RL-trained Finance teacher. Merge initialization can start below SFT warm-up yet finish higher after MOPD. Here, $\lambda$ denotes the global merge scale, defined as the sum of the merge coefficients. IM-MOPD further improves performance in a single run, without an additional training stage or a separate coefficient search. Open and filled circles denote initial and final normalized scores, respectively (reference: 0; teacher: 1). Experimental details are provided in \appautoref{app:initialization_comparison}.}
    \label{fig:main}
\end{figure*}

Building a unified language model requires integrating capabilities acquired through different data and training procedures. Different domains often benefit from different post-training paradigms and recipes, including supervised fine-tuning (SFT), reinforcement learning (RL), or their combinations, with domain-specific choices of data and optimization~\citep{zhang2024scaling, grattafiori2024llama, chu2025sft, blakeman2025nemonano}. These experts may be developed across different timelines, motivating a modular workflow of independent specialization and subsequent integration. Multi-teacher on-policy distillation (MOPD) supports this workflow by transferring the capabilities of domain teachers into a single student. Each teacher provides token-level supervision at the prefixes generated by the student. Recent work demonstrates the practical value of this approach for combining independently developed capabilities~\citep{ma2026mopd, blakeman2026nemotron}.
\looseness=-1

However, it remains unclear whether MOPD can effectively integrate teachers with different post-training histories. Recent work suggests that MOPD can struggle to transfer teacher capabilities when the teacher and student have substantially different SFT histories even though they share the same reference model~\citep{li2026rethinking, blakeman2026nemotron, zhu2026many}. This is particularly important for MOPD, where teacher supervision is provided at student-generated prefixes. Because MOPD distills on student-generated prefixes, the initialization determines the states on which distillation begins and can therefore affect subsequent learning. One way to mitigate this issue is to apply a light SFT warm-up before MOPD~\citep{blakeman2025nemonano}. Such warm-up can bring the student closer to the teachers. However, it still requires designing mixed-domain SFT, including the data mixture and training schedule. This motivates a way to prepare the student for MOPD while preserving the benefits of independent teacher development.
\looseness=-1

We show that merge initialization provides such an alternative. Following the notation of task arithmetic~\citep{ilharco2022editing}, we represent the weight difference between each teacher and the shared reference model as a task vector and initialize the student by adding a weighted combination of these vectors to the reference model.
SFT warm-up incurs additional training cost and requires balancing multiple domain objectives again, weakening the benefit of developing multi-teachers independently. Merge initialization reduces this preparation to selecting and combining teacher task vectors, improving subsequent MOPD without an additional stage of mixed-domain training. Interestingly, initial benchmark performance does not reliably identify a good MOPD initialization. In \figautoref{fig:main}, SFT warm-up starts above merge initialization but finishes below it after MOPD. We further find that effective merging depends on both domain-wise ratios and the global merge scale $\lambda=\sum_i\alpha_i$. In some settings, $\lambda>1$ improves post-MOPD recovery, placing effective initializations outside the simplex of convex parameter averaging. Selecting a good merge initialization therefore requires considering both domain-wise ratios and the global scale, while its quality cannot be reliably assessed from initial performance alone.
\looseness=-1

Rather than directly searching over continuous merge coefficients, we adapt teacher contributions during training. Starting from a uniform merge initialization, we alternate MOPD with small task-vector corrections for under-recovered domains. This replaces a domain-wise continuous coefficient search with repeated merge-or-not decisions using a shared correction size, allowing both the relative domain weights and total task-vector contribution to evolve with the student.
Teacher-continuation analysis (\secautoref{sec:analysis}) shows that teachers perform better with prefixes generated by the merged student than with those from the SFT warm-up student, even when the merged student itself performs similarly or worse. This suggests that the merged student can benefit more from teacher supervision during MOPD. Empirically, our method achieves normalized recovery scores of 86.9 on 4B and 76.0 on 1.7B, compared with 59.0 and 46.5 for SFT warm-up, respectively.
\looseness=-1

\begin{figure*}[t]
    \centering
    \includegraphics[width=\linewidth]{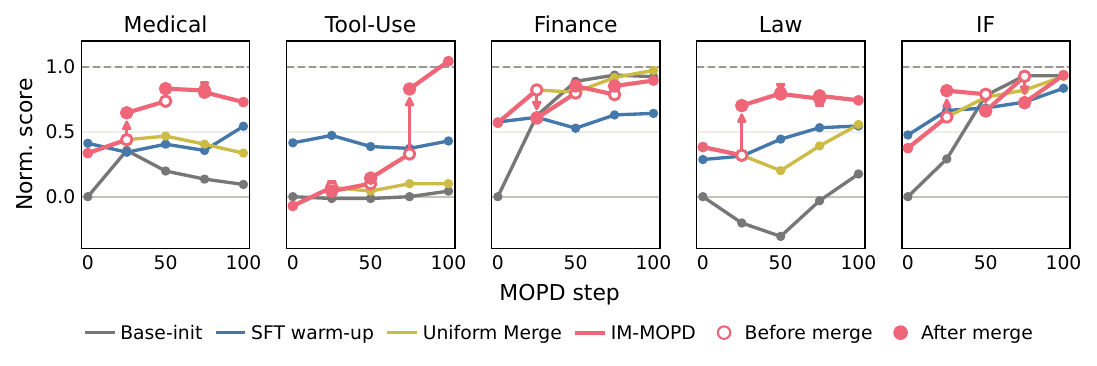}
    \vspace{-2em}
    \caption{\textbf{Iterative merging improves recovery in domains left behind by fixed initialization.}
    On Qwen3-4B, IM-MOPD starts from Uniform Merge and progressively improves under-recovered domains during MOPD. The largest gains are observed in Medical, Tool Use, and Law, while performance in already well-recovered domains remains broadly stable.
    \looseness=-1
    }
    \label{fig:progress}
\end{figure*}

\paragraph{Contributions.}
In summary, our key contributions in this paper are as follows.
\vspace{-5pt}
\begin{itemize}[leftmargin=*, itemsep=2pt]

    \item \textbf{Initialization quality beyond initial performance.} We show that higher initial performance does not necessarily lead to higher post-MOPD performance, and provide additional prefix-level evidence that benchmark scores do not fully characterize a good MOPD initialization (\figautoref{fig:main}, \figautoref{fig:prefix_teacher}).
    \looseness=-1

    \item \textbf{Merge initialization without additional SFT.}
    We show that merging teacher task vectors improves capability recovery after MOPD over base initialization and can outperform the evaluated SFT warm-up baseline. This initialization requires neither an additional SFT stage nor access to the experts' training examples (\figautoref{fig:main}, \tabautoref{tab:main_results}).
    \looseness=-1

    \item \textbf{Iterative merging during MOPD.}
    We propose \textbf{IM-MOPD}, which periodically adds fixed task-vector increments for domains whose recovery remains below a threshold. In a 5-domain setting, IM-MOPD achieves higher average normalized recovery than MOPD baselines using SFT warm-up or uniform merge initialization (\tabautoref{tab:main_results}, \figautoref{fig:progress}).
    \looseness=-1

\end{itemize}

\section{Preliminaries}
\label{sec:preliminaries}

We consider a same-origin multi-teacher setting with $K$ domain teacher policies
$\{\pi_{\phi_i}\}_{i=1}^{K}$, parameterized by
$\{\phi_i\}_{i=1}^{K}$, and their corresponding prompt distributions
$\{\mathcal{D}_i\}_{i=1}^{K}$.
The student policy $\pi_\theta$, parameterized by $\theta$, and all domain
teachers share the same architecture and originate from a common reference
checkpoint $\theta_{\mathrm{ref}}$, after which the teachers are subsequently specialized through different domain-specific post-training procedures.
Our goal is to integrate the capabilities of these independently specialized
teachers into a single student policy $\pi_\theta$.
\looseness=-1

\begin{figure*}[t]
    \centering
    \includegraphics[width=\linewidth]{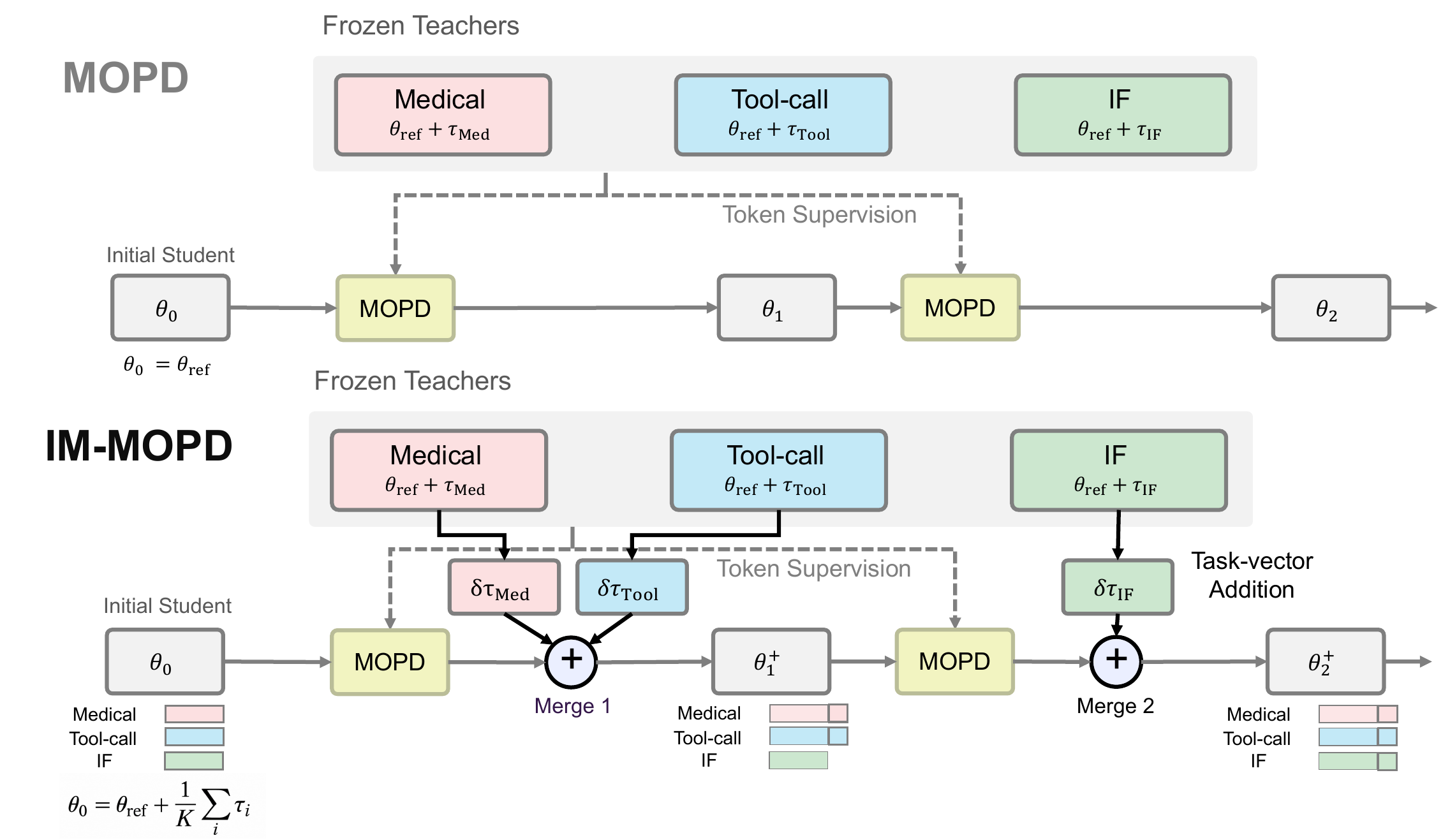}
    \caption{\textbf{Overview of IM-MOPD.}
From a general SFT model $\theta_{\mathrm{ref}}$ and $K$ domain teachers, IM-MOPD starts from a uniform merge of task vectors $\tau_i=\phi_i-\theta_{\mathrm{ref}}$ and alternates MOPD with selective merging: domains below the recovery threshold $\gamma$ receive a fixed increment $\delta\sum_{i\in\mathcal{M}_t}\tau_i$ before the next round of distillation as a direct weight-space correction for an enhanced initialization. For specific details, see \algautoref{alg:immopd}.}
    \label{fig:pipeline}
\end{figure*}

\paragraph{Multi-Teacher On-Policy Distillation (MOPD).}
On-policy distillation trains a student on its own generated responses,
with a teacher providing supervision at the prefixes visited by the
student~\citep{gu2024minillm, agarwal2024policy}.
MOPD extends this procedure to multiple domain teachers by routing each
prompt to its corresponding teacher~\citep{ma2026mopd}.
For a prompt $x\sim\mathcal{D}_i$, the student generates a response
$y=(y_1,\ldots,y_T)\sim\pi_\theta(\cdot\mid x)$.
At token position $t$, the student-generated prefix is
$s_t=(x,y_{<t})$.
MOPD minimizes the token-averaged reverse KL between the student and
the corresponding teacher over these student-generated trajectories:
\looseness=-1
\begin{equation}
    \mathcal{L}_{\mathrm{MOPD}}(\theta)
    =
    \frac{1}{K}\sum_{i=1}^{K}
    \mathbb{E}_{\substack{
        x\sim\mathcal{D}_i\\
        y\sim\pi_\theta(\cdot\mid x)
    }}
    \left[
        \frac{1}{T}
        \sum_{t=1}^{T}
        D_{\mathrm{KL}}\!\left(
            \pi_\theta(\cdot\mid s_t)
            \,\|\,\pi_{\phi_i}(\cdot\mid s_t)
        \right)
    \right].
    \label{eq:mopd_objective}
\end{equation}
In practice, we optimize this objective using the policy-gradient formulation of on-policy distillation for computational efficiency; implementation details are provided in \appautoref{app:mopd_training}.
Because supervision is conditioned on student-generated prefixes, initialization shapes the prefixes encountered during distillation and thus the subsequent MOPD trajectory.
\looseness=-1

\vspace{-10pt}
\paragraph{Model Merging.}
Model merging combines compatible model checkpoints directly in parameter space without additional training~\citep{izmailov2018averaging, wortsman2022model}. In our setting, all teachers share the reference checkpoint $\theta_{\mathrm{ref}}$, allowing each teacher to be represented by a task vector
$\tau_i=\phi_i-\theta_{\mathrm{ref}}$~\citep{ilharco2022editing}.
We parameterize a merged model as
$\theta_{\mathrm{merge}}(\mathbf{\alpha})
=\theta_{\mathrm{ref}}+\sum_{i=1}^{K}\alpha_i\tau_i$,
where $\alpha_i\geq0$ controls the contribution of teacher $i$.
We define the global merge scale as $\lambda=\sum_i\alpha_i$.
Standard weighted parameter averaging corresponds to $\lambda=1$, with uniform averaging given by $\alpha_i=1/K$.
We do not otherwise constrain $\lambda$ to one, allowing configurations with larger total teacher contribution.
A merged model can be used as the initial MOPD student,
$\theta_0=\theta_{\mathrm{merge}}(\mathbf{\alpha})$.
\looseness=-1

\section{Method}
\label{sec:method}

\begin{figure*}[t]
    \centering
    \captionsetup[subfigure]{justification=centering,singlelinecheck=true}
    \begin{subfigure}[t]{0.29\textwidth}
        \centering
        \begin{minipage}[t][1.75in][b]{\linewidth}
            \vspace{0pt}
            \includegraphics[width=\linewidth]{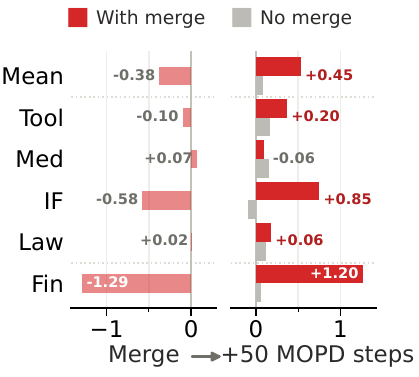}
        \end{minipage}
        \makebox[\linewidth][l]{\hspace*{\dimexpr3pt+2mm\relax}\parbox[t]{\linewidth}{\vspace{-1pt}\caption{5-domain merge effect.}\label{fig:obs_mid_merge}}}
    \end{subfigure}\hfill
    \begin{subfigure}[t]{0.29\textwidth}
        \centering
        \begin{minipage}[t][1.75in][b]{\linewidth}
            \vspace{0pt}
            \includegraphics[width=\linewidth]{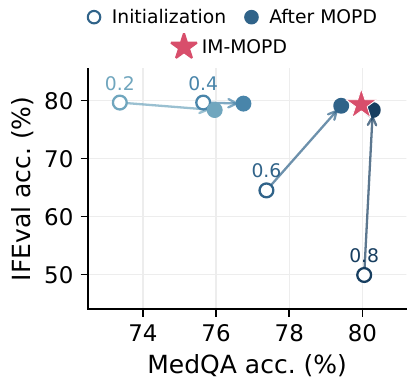}
        \end{minipage}
        \makebox[\linewidth][l]{\hspace*{\dimexpr3pt+2mm\relax}\parbox[t]{\linewidth}{\vspace{-1pt}\caption{2-domain grid search.}\label{fig:obs_2domain}}}
    \end{subfigure}\hfill
    \begin{subfigure}[t]{0.40\textwidth}
        \centering
        \begin{minipage}[t][1.75in][b]{\linewidth}
            \vspace{0pt}
            \includegraphics[width=\linewidth]{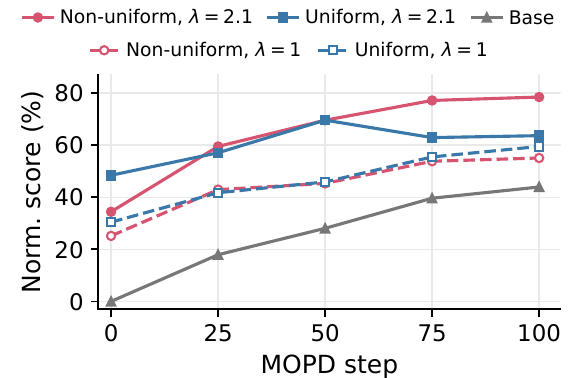}
        \end{minipage}
        \makebox[\linewidth][l]{\hspace*{\dimexpr3pt+2mm\relax}\parbox[t]{\linewidth}{\vspace{-1pt}\caption{Merge ratios and scale.}\label{fig:obs_5domain}}}
    \end{subfigure}
    \caption{\textbf{Selecting a fixed merge initialization requires accounting for subsequent learning, ratios, and scale.}
    \textbf{(a)} From the same 5-domain MOPD checkpoint, we continue training with or without a merge intervention. Although merging can reduce performance immediately, the intervened branch achieves higher recovery after 50 additional MOPD updates.
    \textbf{(b)} 2-domain Medical-IF coefficient grid vs. IM-MOPD. Open/filled circles denote performance before/after MOPD for fixed merge initializations; numbers indicate the Medical coefficient, with the remainder assigned to IF, and the star denotes IM-MOPD.
    \textbf{(c)} 5-domain MOPD under different merge initialization. Both ratios and scale affect post-MOPD recovery.}
    \label{fig:main_analysis}
\end{figure*}

\paragraph{Overview.}
We study how student initialization influences the integration of domain teachers with heterogeneous post-training histories under MOPD. All teachers originate from a shared reference model $\theta_{\mathrm{ref}}$ but are independently specialized through SFT or RLVR. Details of teacher construction and the experimental setup are provided in \appautoref{app:teacher_training} and~\appautoref{app:mopd_training}. Following \citet{ma2026mopd}, let $s_i(\theta)$ denote the evaluation
performance of model $\theta$ on domain $i$.
We define the normalized domain score as
$\tilde{s}_i(\theta)=
\frac{s_i(\theta)-s_i(\theta_{\mathrm{ref}})}
{s_i(\phi_i)-s_i(\theta_{\mathrm{ref}})}$,
such that the shared reference model has score $0$ and the corresponding
domain expert has score $1$.
We also report the average normalized score
$\tilde{s}(\theta)=
\frac{1}{K}\sum_{i=1}^{K}\tilde{s}_i(\theta)$.
\looseness=-1

\subsection{Observations}
\label{subsec:observations}

\paragraph{Merge initialization provides an effective starting point for MOPD.}
As shown in \figautoref{fig:main}, MOPD from the shared reference model struggles to recover the capabilities of SFT-trained teachers such as Medical and Tool-Use, while approaching teacher-level performance on Finance. This uneven recovery suggests that the choice of student initialization can affect how effectively different teacher capabilities are transferred. SFT warm-up partially alleviates this issue, but introduces an additional training cost and requires separate choices of data mixture and training schedule. We find that initializing the student by merging teacher task vectors substantially improves subsequent MOPD recovery without any additional training. This suggests that model merging can serve as a simple and low-cost way to prepare students for MOPD.
\looseness=-1

\vspace{-10pt}
\paragraph{A stronger student is not necessarily a better MOPD initialization.}
In \figautoref{fig:main}, SFT warm-up yields higher initial performance than the scaled merge initialization ($\lambda=2.1$) on Tool-Use, yet the ordering reverses after MOPD. IM-MOPD further improves recovery despite not starting from the strongest benchmark performance. A similar pattern appears for merge interventions during training. In \figautoref{fig:obs_mid_merge}, we compare continued MOPD with and without intermediate merge interventions from the same checkpoint. Although a merge intervention can temporarily reduce normalized performance, the intervened branch achieves higher performance after 50 additional MOPD updates. These results indicate that immediate benchmark performance alone is insufficient for selecting states for subsequent MOPD.
\looseness=-1

\vspace{-10pt}
\paragraph{Selecting good merge coefficients becomes increasingly difficult as the number of domains grows.}
Even with two domains, the coefficient sweep in \figautoref{fig:obs_2domain} yields different post-MOPD trade-offs, and the best initialization by average normalized score is not the best after MOPD. Evaluating a candidate therefore requires considering the learning it enables. With 5-domains, \figautoref{fig:obs_5domain} shows that both merge ratios and scale matter: non-uniform weighting improves recovery at the larger scale, while normalizing the coefficients to sum to one weakens the result. Increasing the uniform scale helps but does not match the stronger non-uniform configuration. Good coefficients thus require jointly choosing domain-wise ratios and total contribution, with each candidate assessed through MOPD. As the teacher set grows, this makes a fixed coefficient search increasingly costly, motivating progressive merge-or-not decisions during training.
\looseness=-1

\subsection{Iterative Merging for MOPD (IM-MOPD)}
\label{subsec:immopd}

\begin{wrapfigure}{r}{0.43\columnwidth}
\vspace{-0.6\baselineskip}

\footnotesize

\hrule
\vspace{0.35em}

\refstepcounter{algorithm}
\noindent
\textbf{Algorithm \thealgorithm }
IM-MOPD
\label{alg:immopd}

\vspace{0.35em}
\hrule
\vspace{0.4em}
\algrenewcommand\algorithmicindent{1.2em}
\begin{algorithmic}

\Statex \textbf{Require:}
$\theta_{\mathrm{ref}}$ reference model;
$\{\phi_i\}_{i=1}^{K}$ teachers;
$\tau_i=\phi_i-\theta_{\mathrm{ref}}$ task vectors;
$n$ MOPD iterations;
$H$ correction interval;
$\delta$ fixed correction scale;
$\gamma$ threshold

\State $\theta \gets
\theta_{\mathrm{ref}}
+\frac{1}{K}\sum_{i=1}^{K}\tau_i$

\For{$t=1,\ldots,n$}
    \State $\theta \gets
    \textsc{MOPD}(\theta;\{\phi_i\}_{i=1}^{K})$
    \If{$t\bmod H=0$ \textbf{and} $t<n$}
        \State $\mathcal M \gets
        \{i \mid \tilde{r_i}(\theta)\leq\gamma\}$
        \State $\theta \gets
        \theta+\delta\sum_{i\in\mathcal M}\tau_i$
    \EndIf
\EndFor

\State \Return $\theta$

\end{algorithmic}
\vspace{0.35em}
\hrule
\vspace{-0.3\baselineskip}
\end{wrapfigure}

Motivated by the observations above, we replace one-shot merge coefficient search with an iterative procedure that interleaves MOPD updates with small, fixed-scale additions of selected teacher task vectors (\figautoref{fig:pipeline}). IM-MOPD initializes the student with a uniform merge, $\theta_0=\theta_{\mathrm{ref}}+\frac{1}{K}\sum_{i=1}^{K}\tau_i$, and alternates distillation with selective task-vector corrections for domains whose capabilities remain insufficiently recovered. This allows the effective merge weights to adapt throughout training without determining the full coefficient set in advance.
\looseness=-1

At intervals of $H$ MOPD updates, we assess domain recovery on a validation set. Let $\tilde{r_i}(\theta)$ denote the normalized score evaluated on validation data for domain $i$, with the reference model and corresponding teacher anchoring zero and one. We select domains whose recovery is at or below a threshold $\gamma$, forming $\mathcal{M}=\{i\mid \tilde{r_i}(\theta)\leq\gamma\}$, and update the student as $\theta\leftarrow\theta+\delta\sum_{i\in\mathcal{M}}\tau_i$. Here, $\delta>0$ is a fixed correction scale shared across domains and intervention points. \algautoref{alg:immopd} summarizes the procedure.
\looseness=-1

MOPD resumes from the corrected student, and subsequent merge decisions reflect the recovery achieved through further learning. A domain may receive additional corrections if its recovery remains below the threshold.
These task-vector additions allow both the relative merge ratios and the total merge scale to evolve during training, without constraining the cumulative coefficients to sum to one. By interleaving these corrections with distillation, IM-MOPD progressively shapes the student from which further teacher-guided learning proceeds.
\looseness=-1

\section{Experiments}
\label{sec:experiments}

\subsection{Experimental Setup}
\label{subsec:experimental_setup}

\paragraph{Reference model and teachers.}
We first fine-tune Qwen3-4B-Base and Qwen3-1.7B-Base~\citep{yang2025qwen3} on OpenThoughts3~\citep{guha2026openthoughts} to obtain a common reference checkpoint for the student and domain teachers at each model size. We then train 5 domain teachers separately from this reference: Medical, Law, and Tool Use through SFT, and Finance and Instruction Following (IF) through reinforcement learning with verifiable rewards (RLVR). For detailed training and data configurations, please refer to \appautoref{app:teacher_training}.
\looseness=-1

\vspace{-10pt}
\paragraph{Baselines.}
We compare our method with three representative initialization baselines. \emph{MOPD} starts from the common reference model~\citep{ma2026mopd}. \emph{Uniform Merge} starts from the equal-weight average of the five teachers and serves as the static merge initialization baseline, isolating the effect of iterative adaptation from the choice of a more sophisticated one-shot merging rule. \emph{SFT Warm-up} first trains the student on mixed teacher-generated samples before MOPD. More details on the baselines are provided in \appautoref{app:mopd_training}--\ref{app:merge_details}.
\looseness=-1

\vspace{-10pt}
\paragraph{Iterative-merging configurations.}
Both model sizes start from uniform merging and use the same validation-based recovery rule. We set the recovery threshold to $\gamma=0.6$ for 4B and $\gamma=0.4$ for 1.7B, the fixed correction scale to $\delta=0.3$ for both, and the correction interval to $H=25$. Every $H$ MOPD updates, we add the corresponding teacher task vector for domains whose recovery is at or below $\gamma$. \appautoref{app:merge_details} provides the validation protocol and correction schedules.
\looseness=-1

\vspace{-10pt}
\paragraph{Evaluations.}
We evaluate Medical on MedQA~\citep{jin2021disease}, Law on CaseHOLD~\citep{zheng2021does}, Finance on FinQA~\citep{chen2021finqa}, Instruction Following on IFBench~\citep{pyatkin2025generalizing}, and Tool Use on $\tau^2$-Telecom~\citep{barres2025tau}. We report answer accuracy for MedQA and CaseHOLD, numeric-answer accuracy for FinQA, prompt-level loose accuracy for IFBench, and task success for telecom tasks. For the main results, we report pass@1 for Tool Use and avg@3 for the other benchmarks, together with the average normalized score across the five domains. Evaluation protocols are provided in \appautoref{app:eval_details}.
\looseness=-1

\subsection{Main Results}
\label{subsec:main_results}

\begin{table*}[!t]
\caption{
\textbf{MOPD results across 5-domains at Qwen3-4B and Qwen3-1.7B.} IM-MOPD achieves the highest average normalized score among the compared student initialization methods at both model sizes, without an additional SFT stage. Domain Teacher reports the performance of the domain specialist for each domain. The best result in each column within each model size is shown in \textbf{bold}.
\looseness=-1
}
\label{tab:main_results}
\centering

\begingroup
\small
\setlength{\tabcolsep}{6pt}
\renewcommand{\arraystretch}{1.08}

\begin{tabular}{lrrrrrr}
\toprule
\textbf{Method} & \textbf{MedQA} & \textbf{CaseHOLD} & \textbf{FinQA} & \textbf{IFBench} & $\boldsymbol{\tau^2}$ & \textbf{Norm.} \\
\midrule

\rowcolor{blue!8}
\multicolumn{7}{c}{\footnotesize \textbf{Qwen3-4B}} \\
\midrule

Qwen3-4B-OT3 & 69.5 & 61.8 & 59.1 & 24.9 & \,\,\,5.3 & \,\,\,0.0 \\
Domain Teacher & 80.7 & 73.4 & 75.0 & 59.4 & 66.7 & 100.0 \\
\midrule

MOPD & 69.1 & 64.7 & 74.3 & 56.9 & \,\,\,7.9 & 42.8 \\
SFT Warm-up + MOPD & 73.9 & 68.2 & 71.1 & 53.4 & 31.6 & 59.0 \\
Uniform Merge + MOPD & 72.7 & 67.6 & 74.7 & 53.6 & 11.4 & 53.8 \\

\rowcolor{gray!15}
\textbf{IM-MOPD} & \textbf{76.6} & \textbf{70.9} & 74.0 & \textbf{57.8} & \textbf{69.3} & \textbf{86.9} \\
\midrule

\rowcolor{blue!8}
\multicolumn{7}{c}{\footnotesize \textbf{Qwen3-1.7B}} \\
\midrule

Qwen3-1.7B-OT3 & 44.1 & 44.6 & 42.4 & 18.4 & \,\,\,7.9 & \,\,\,0.0 \\
Domain Teacher & 58.5 & 69.6 & 62.4 & 46.4 & 28.1 & 100.0 \\
\midrule

MOPD & 49.0 & 52.2 & \textbf{59.1} & 36.8 & \,\,\,0.0 & 34.9 \\
SFT Warm-up + MOPD & 52.7 & 56.6 & 54.8 & 32.6 & 10.5 & 46.5 \\
Uniform Merge + MOPD & 51.2 & \textbf{59.9} & 58.6 & 34.6 & \,\,\,4.4 & 46.3 \\

\rowcolor{gray!15}
\textbf{IM-MOPD} & \textbf{53.7} & 59.4 & 55.9 & \textbf{42.9} & \textbf{28.1} & \textbf{76.0} \\

\bottomrule
\end{tabular}

\endgroup
\vspace{3pt}
\end{table*}

\paragraph{Merge initialization improves recovery, but leaves uneven domain transfer.}
\tabautoref{tab:main_results} shows that starting MOPD from a uniform merge substantially improves aggregate recovery over starting from the reference model, without requiring an additional SFT stage. SFT warm-up provides a further improvement in aggregate recovery, but requires an additional training stage. Despite these gains, both approaches leave some domains substantially under-recovered, most notably Tool Use. Thus, a better initialization improves MOPD overall, but does not by itself ensure balanced capability integration across domains.
\looseness=-1

\vspace{-10pt}
\paragraph{Iterative merging improves aggregate capability integration.}
For Qwen3-4B, iterative merging increases the average normalized recovery from 53.8\% with uniform initialization to 86.9\%. For Qwen3-1.7B, IM-MOPD reaches 76.0\%, compared with 46.5\% for SFT warm-up, and recovers Tool Use to the teacher level. Across both model sizes, IM-MOPD achieves higher scores than SFT warm-up on all five benchmarks. Relative to uniform initialization, the largest improvements are observed in Medical and Tool Use, while performance on Finance and Instruction Following remains broadly comparable. These results show that iterative corrections can improve recovery in domains that remain under-recovered after initialization without substantially degrading domains already transferred effectively by MOPD.
\looseness=-1

\subsection{Ablations}
\label{subsec:ablations}

\begin{table}[ht]
\caption{\textbf{Merge-timing ablation on Qwen3-4B.} We take the final cumulative merge coefficients produced by IM-MOPD and apply them either entirely at initialization or entirely after MOPD. Both controls underperform the original IM-MOPD schedule, showing that when the same final merge coefficients are applied matters. The best result in each column is shown in \textbf{bold}.
\looseness=-1
}
\label{tab:merge_ablations}
\centering

\begingroup
\small
\begin{tabular*}{0.9\textwidth}{@{\extracolsep{\fill}}lrrrrrr@{}}
\toprule
\textbf{Merge timing} & \textbf{MedQA} & \textbf{CaseHOLD} & \textbf{FinQA} & \textbf{IFBench} & $\boldsymbol{\tau^2}$ & \textbf{Norm.} \\
\midrule
No merge (Base MOPD) & 69.1 & 64.7 & \textbf{74.3} & 56.9 & \,\,\,7.9 & 42.8 \\
\midrule
All at initialization & \textbf{76.3} & 70.8 & 74.0 & 52.4 & 57.0 & 79.0 \\
All after MOPD & 68.1 & \textbf{71.0} & 59.6 & 38.2 & 10.5 & 23.3 \\

% \rowcolor{gray!15}
Iterative (IM-MOPD) & \textbf{76.6} & 70.9 & 74.0 & \textbf{57.8} & \textbf{69.3} & \textbf{86.9} \\
\bottomrule
\end{tabular*}
\endgroup
\end{table}

\paragraph{Iterative merging versus merging only at initialization.}
We compare IM-MOPD with an initialization-only variant that merges teacher task vectors using the same cumulative coefficients, followed by the same number of MOPD updates (\tabautoref{tab:merge_ablations}). Iterative merging reaches 86.9\% normalized recovery, versus 79.0\% for the initialization-only variant, with the largest gain in Tool Use. With the cumulative coefficients held fixed, distributing task-vector additions across distillation is more effective than applying them all at the start.
\looseness=-1

\vspace{-10pt}
\paragraph{Merging and MOPD reinforce each other.}
After MOPD from the reference model with no additional initialization, we apply the final cumulative merge coefficients obtained by IM-MOPD to the trained student. As shown in \tabautoref{tab:merge_ablations}, this late merging achieves only 23.3\% normalized recovery, below Base MOPD and far below IM-MOPD. Together with the initialization-only control, this result shows that the gains of IM-MOPD cannot be explained simply by the final task-vector displacement induced by merging. The same merge coefficients are substantially more effective when interleaved with MOPD, indicating that subsequent teacher-guided updates after each correction are important for the final recovery.
\looseness=-1

\Needspace{14\baselineskip}
\begin{wraptable}[10]{r}{0.51\textwidth}
\vspace{-\intextsep}
\centering
\caption{\textbf{IM-MOPD sensitivity to $\gamma$ and $\delta$.} Average normalized recovery (\%) across 5 domains.}
\label{tab:ramp_ablation}
\label{tab:threshold_ablation}
\small
\setlength{\tabcolsep}{3pt}
\begin{tabular*}{\linewidth}{@{\extracolsep{\fill}}lrrrr@{}}
\toprule
Model & $\gamma$ & $\delta=.2$ & $\delta=.3$ & $\delta=.4$ \\
\midrule
Qwen3-4B & 0.6 & 80.3 & 86.9 & 86.1 \\
\bottomrule
\end{tabular*}
\par\vspace{5pt}
\begin{tabular*}{\linewidth}{@{\extracolsep{\fill}}lrrrr@{}}
\toprule
Model & $\gamma$ & $\delta=.1$ & $\delta=.2$ & $\delta=.3$ \\
\midrule
Qwen3-1.7B & 0.4 & 62.2 & 57.2 & 76.0 \\
 & 0.6 & 57.7 & 65.0 & 69.8 \\
\bottomrule
\end{tabular*}
\vspace{-4pt}
\end{wraptable}

\paragraph{Hyperparameter sensitivity.}
We evaluate sensitivity to the recovery threshold $\gamma\in\{0.4,0.6\}$ and correction size $\delta\in\{0.1,0.2,0.3,0.4\}$ in \tabautoref{tab:ramp_ablation}. Qwen3-4B maintains over 80\% normalized recovery across tested correction sizes, while Qwen3-1.7B performs best with $\gamma=0.4$ and $\delta=0.3$. Effective settings can thus be identified from a small candidate set. More detailed results are provided in \appautoref{app:ablation_results}.
\looseness=-1

\section{Analysis}
\label{sec:analysis}

\begin{figure}[!ht]
    \centering
    \begin{subfigure}[t]{0.485\linewidth}
        \centering
        \includegraphics[width=\linewidth]{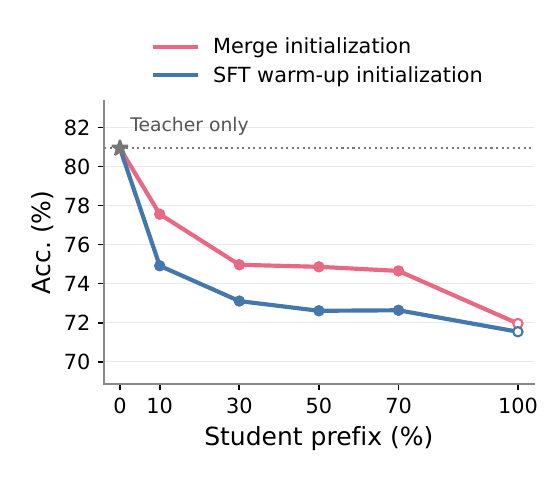}
        \caption{Medical (MedQA).}
    \end{subfigure}
    \hfill
    \begin{subfigure}[t]{0.485\linewidth}
        \centering
        \includegraphics[width=\linewidth]{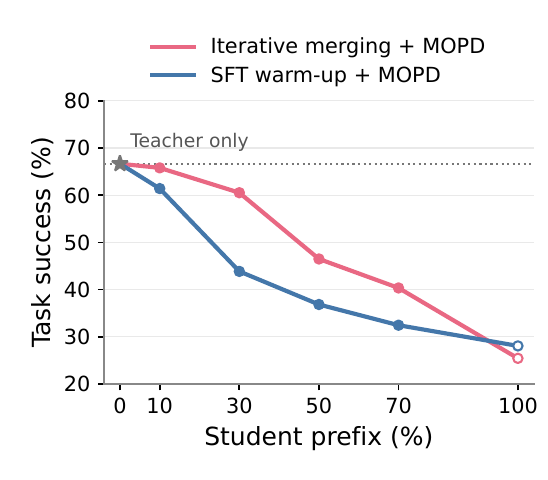}
        \caption{Tool Use ($\tau^2$-Telecom).}
    \end{subfigure}
    \caption{\textbf{Teacher continuation is stronger from merge-based student prefixes on Qwen3-4B.} In Medical, teacher continuation is consistently stronger from merge-initialized prefixes than from SFT-warm-up prefixes despite similar student-only performance. In Tool Use, IM-MOPD yields higher continuation success from short prefixes despite lower student-only performance at the evaluated checkpoint. The 0\% and 100\% endpoints denote teacher-only and student-only performance, respectively.
    \looseness=-1
    }
    \label{fig:prefix_teacher}
\end{figure}

\paragraph{Teacher continuation from student-generated prefixes.}
Prior work evaluates the reliability of teacher supervision in OPD by continuing from student-generated prefixes~\citep{li2026rethinking}. We adopt this diagnostic to assess teacher--student compatibility for merged and SFT warm-up students. Each student generates an initial prefix, after which the corresponding domain teacher completes the response. We vary the fraction generated by the student and evaluate task performance (\figautoref{fig:prefix_teacher}). Higher continuation performance indicates greater compatibility between the student-generated prefix and the teacher's subsequent generation. Protocol details are provided in \appautoref{app:prefix_teacher}.
\looseness=-1

Across Medical and Tool Use, teacher continuation performance drops even with short prefixes from the SFT warm-up student, whereas prefixes from the merged student better preserve it over the early portion of the trajectory (\figautoref{fig:prefix_teacher}). These results are consistent with prior findings that teacher--student compatibility matters for effective OPD~\citep{li2026rethinking}. This advantage persists even when the merged student's standalone performance is similar to or lower than that of the SFT warm-up student. Together, these results suggest that merging helps the student generate prefixes from which teachers can more successfully continue, allowing the student to benefit more from teacher supervision during MOPD.
\looseness=-1

\section{Related Work}
\label{sec:related_work}

\paragraph{On-Policy Distillation for Large Language Models (LLMs).}
Knowledge distillation (KD) traditionally transfers teacher knowledge by matching predictive distributions on a fixed training distribution~\citep{hinton2015distilling}. \citealp{kim2016sequence} extended the paradigm to apply sequence-level KD from teacher-generated sequences. However, such offline supervision exposes the student primarily to teacher- or data-induced prefixes, creating a mismatch between train and inference setups. On-policy distillation (OPD) addresses this mismatch by instead querying the teacher on student-generated trajectories \citep{agarwal2024policy}, and is commonly paired with reverse-KL objectives that naturally optimize the student distribution over its own visited states \citep{gu2024minillm}. This formulation has become increasingly relevant for reasoning LLMs, where the context has scaled with the reasoning length \citep{xu2026deepseekv4,blakeman2026nemotron,ma2026mopd,team2026kimik3}. Yet, the benefit of on-policy supervision depends on the teacher remaining informative on states visited by the student: as teacher--student distributions diverge, teacher feedback can become less effective or unreliable \citep{xu2025speculative,li2026rethinking, blakeman2026nemotron}. Recent methods therefore selectively constrain or adapt distillation according to teacher--student alignment \citep{xing2026trust}, highlighting distributional compatibility as a central challenge in effective OPD.
\looseness=-1

\vspace{-10pt}
\paragraph{Consolidating Domain Specialists.}
As post-training increasingly targets heterogeneous capabilities often through reinforcement learning (RL), consolidating these divergent capabilities into a single model has become a central challenge. Existing approaches differ in how capabilities are acquired and integrated. Mixed RL jointly optimizes a shared policy over a mixture of domains, enabling direct capability integration but coupling domains with potentially heterogeneous data, rewards, rollout lengths, and optimization dynamics \citep{blakeman2025nemonano}. Cascade RL instead applies domain-specific RL stages sequentially, allowing each stage to use domain-specific environments and optimization configurations \citep{wang2025nemotroncascade1}. While carefully designed cascades can largely preserve previously acquired capabilities, capability drift can accumulate across increasingly specialized stages, motivating intermediate consolidation or stabilization \citep{yang2026nemocascade2}. Alternatively, domain specialists can be trained independently and subsequently consolidated through parameter merging \citep{team2025introducing} or multi-teacher on-policy distillation (MOPD) \citep{ma2026mopd}. Our work builds on this specialist-consolidation paradigm and investigates whether iterative merging can progressively integrate heterogeneous capabilities while maintaining effective teacher--student alignment.
\looseness=-1

\section{Conclusion}
\label{sec:conclusion}

We study MOPD when domain teachers share a common reference model but undergo different post-training procedures. We find that initial benchmark performance is an unreliable criterion for selecting MOPD initializations, and that effective merge initialization depends on both relative teacher weights and the overall merge scale. Motivated by these observations, we introduce IM-MOPD, which starts from a uniform merge and progressively adds task-vector corrections for domains that remain under-recovered during distillation. Across five domains at both 4B and 1.7B scales, IM-MOPD achieves higher aggregate capability recovery than uniform merging, and SFT warm-up, without requiring an additional SFT stage. Merge-timing ablations further show that applying the same final cumulative merge coefficients entirely before or after MOPD is less effective than interleaving the corresponding corrections with distillation. Together, these results suggest that teacher contributions are more effectively determined progressively during MOPD than fixed entirely before training.
\looseness=-1

\section*{Acknowledgements}
This work was conducted in collaboration with and supported by Kakao Corp.
% This study uses existing benchmark data and simulated tool-use environments. Distillation and merging can transfer factual errors, biases, and unsafe behavior from teacher models. Our medical, legal, and financial evaluations measure benchmark performance and do not establish readiness for professional decision-making. Applying these models in such settings or to real-world tools requires separate safety assessment and appropriate human oversight.
% \looseness=-1

\section*{Reproducibility Statement}
\algautoref{alg:immopd} specifies IM-MOPD. \appautoref{app:teacher_training} documents teacher training and data sources, while Appendices~\ref{app:mopd_training}--\ref{app:warmup_details} describe MOPD and SFT warm-up settings. Validation data and merge schedules are given in \appautoref{app:merge_details}, and evaluation protocols in \appautoref{app:eval_details}. Appendices~\ref{app:initialization_comparison}--\ref{app:two_domain} provide additional ablation results, the student-prefix diagnostic procedure, and two-domain results.
\looseness=-1

\clearpage
\bibliographystyle{plainnat}
\bibliography{paper}

\clearpage
\appendix{\setlength{\cftbeforesecskip}{16pt}
\setlength{\cftbeforesubsecskip}{4pt}
\renewcommand\cftsecpagefont{\color{RoyalBlue}}
\renewcommand\cftsubsecpagefont{\color{RoyalBlue}}
\renewcommand\cftsubsubsecpagefont{\color{RoyalBlue}}
\renewcommand{\contentsname}{\large{Contents}}
{
  \hypersetup{linkcolor=}
  \tableofcontents
}

\clearpage

% ==================================================
%                   Next Section
% ==================================================
\section{Experimental Details}
\label{app:experimental_details}

Training was performed on NVIDIA A100 80GB GPUs.

\subsection{Teacher Training}
\label{app:teacher_training}

We first fine-tune Qwen3-4B-Base\footnote{\url{https://huggingface.co/Qwen/Qwen3-4B-Base}} and Qwen3-1.7B-Base\footnote{\url{https://huggingface.co/Qwen/Qwen3-1.7B-Base}} on OpenThoughts3\footnote{\url{https://huggingface.co/datasets/open-thoughts/OpenThoughts3-1.2M}} to obtain a common reference at each model size. Domain teachers are initialized from the corresponding reference. \tabautoref{tab:teacher_recipes} and~\tabautoref{tab:teacher_recipes_17} summarize their training configurations. For GRPO, batch size is given as prompts $\times$ completions.
\looseness=-1

\begin{table}[ht]
\centering
\caption{Training configurations of the 4B domain teachers.}
\label{tab:teacher_recipes}
\small
\setlength{\tabcolsep}{4pt}
\begin{tabular}{llrrrr}
\toprule
Teacher & Training & Batch & Length & Peak LR & Tokens (M) \\
\midrule
Medical & SFT & 16 & 8,192 & $10^{-5}$ & 426 \\
Law & SFT & 128 & 20,480 & $10^{-5}$ & 269 \\
Tool Use & SFT & 64 & 32,768 & $5\cdot10^{-6}$ & 796 \\
Finance & GRPO & $32\times8$ & 16,384 & $3\cdot10^{-6}$ & 315 \\
IF & GRPO & $64\times8$ & 32,768 & $3\cdot10^{-6}$ & 483 \\
\bottomrule
\end{tabular}
\end{table}

\begin{table}[ht]
\centering
\caption{Training configurations of the 1.7B domain teachers.}
\label{tab:teacher_recipes_17}
\small
\setlength{\tabcolsep}{4pt}
\begin{tabular}{llrrrr}
\toprule
Teacher & Training & Batch & Length & Peak LR & Tokens (M) \\
\midrule
Medical & SFT & 16 & 8,192 & $10^{-5}$ & 426 \\
Law & SFT & 64 & 20,480 & $10^{-5}$ & 269 \\
Tool Use & SFT & 64 & 32,768 & $5\cdot10^{-6}$ & 796 \\
Finance & GRPO & $32\times8$ & 16,384 & $3\cdot10^{-6}$ & 275 \\
IF & GRPO & $64\times8$ & 32,768 & $3\cdot10^{-6}$ & 424 \\
\bottomrule
\end{tabular}
\end{table}

\paragraph{SFT data.}
Medical uses 40,853 correct, deduplicated samples generated by Qwen3.6-35B-A3B from MedQA training prompts. Law uses 56,123 correct, deduplicated samples generated by the same model from CaseHOLD and bar-exam prompts. Tool Use uses 32,758 retail, airline, and telecom samples from \texttt{inclusionAI/AReaL-tau2-data}\footnote{\url{https://huggingface.co/datasets/inclusionAI/AReaL-tau2-data}}, after removing overlength samples and telecom task-ID overlap. Tool schemas are preserved, and supervision covers the final assistant turn with earlier thinking removed.
\looseness=-1

\vspace{-10pt}
\paragraph{SFT optimization.}
SFT uses the Trainer implementation in Hugging Face Transformers\footnote{\url{https://github.com/huggingface/transformers}}. All three domains use packing, zero weight decay, gradient-norm clipping at 1, and cosine decay with 5\% warm-up. Tool Use additionally uses a 10\% learning-rate floor.
\looseness=-1

\vspace{-10pt}
\paragraph{RL data.}
Finance uses 7,565 numeric-answer prompts from the training splits of FinQA and TAT-QA.\footnote{\url{https://github.com/NExTplusplus/TAT-QA}} IF uses 11,728 verifiable instruction-following prompts from the \texttt{rlvr1} blend of Nemotron-RL-Ultra-Training-Blends.\footnote{\href{https://huggingface.co/datasets/nvidia/Nemotron-RL-Ultra-Training-Blends}{\nolinkurl{https://huggingface.co/datasets/nvidia/}\\\nolinkurl{Nemotron-RL-Ultra-Training-Blends}}}
\looseness=-1

\vspace{-10pt}
\paragraph{GRPO.}
Finance and IF use NeMo-RL\footnote{\url{https://github.com/NVIDIA-NeMo/RL}} GRPO with numeric-answer and instruction-constraint rewards, respectively. Both use AdamW with zero weight decay, ten learning-rate warm-up updates, and a constant rate thereafter. Training uses clipped policy updates and overlength penalties, without a reference KL penalty. Sampling uses temperature 1 and top-$p$ 1.
\looseness=-1

\subsection{MOPD Training Configuration}
\label{app:mopd_training}

\paragraph{MOPD data.}
Medical uses 8,384 four- and five-option MedQA training prompts. Law uses 31,563 CaseHOLD and bar-exam prompts.\footnote{Bar-exam data: \url{https://huggingface.co/datasets/reglab/barexam_qa}.} Finance and IF reuse the RL prompt pools described in \appautoref{app:teacher_training}. Tool Use uses 31,718 dialogue contexts from the retail, airline, and telecom portions of AReaL-tau2-data, with each context serving as a prompt for the next assistant turn.
\looseness=-1

\vspace{-10pt}
\paragraph{MOPD optimization.}
All student methods use 100 MOPD updates, a batch of 128 prompts, uniform domain sampling, and one completion per prompt. Policy-gradient distillation clips advantage magnitude at 5 and masks truncated completions, without task-reward or reference-KL terms. The learning rate is $10^{-6}$, and completions are capped at 16,384 tokens. Sampling uses temperature 1 and top-$p$ 1.
\looseness=-1

\subsection{SFT Warm-up Baseline}
\label{app:warmup_details}

The warm-up baseline trains on mixed samples generated by the five teachers at the corresponding model size until the student reaches approximately 30\% normalized recovery. MOPD then starts from this warm-up checkpoint (\tabautoref{tab:warmup_recipes}).
\looseness=-1

\begin{table}[ht]
\centering
\caption{SFT warm-up configurations.}
\label{tab:warmup_recipes}
\small
\setlength{\tabcolsep}{4pt}
\begin{tabular}{llrrrrr}
\toprule
Student & Training & Batch & Length & Peak LR & Tokens (M) \\
\midrule
4B & SFT & 32 & 32,768 & $10^{-5}$ & $167$ \\
1.7B & SFT & 32 & 32,768 & $10^{-5}$ & $171$ \\
\bottomrule
\end{tabular}
\end{table}

Warm-up uses packed sequences and cosine learning-rate decay with 5\% warm-up.
\looseness=-1

\subsection{Main Experiment Settings}
\label{app:merge_details}

Main results use $\gamma=0.6$ for 4B and $\gamma=0.4$ for 1.7B, with $\delta=0.3$ for both. Corrections are applied after updates 25, 50, and 75 according to the recovery rule in \algautoref{alg:immopd}. \tabautoref{tab:merge_schedules} lists the selected domains for each configuration.
\looseness=-1

\paragraph{Validation data.}
For Tool Use, we sample 128 tasks from the full dataset. For each other domain, we sample 256 prompts from a validation set held out from its training pool.
\looseness=-1

\begin{table}[ht]
\centering
\caption{Realized merge schedules. T/M/L/F/I denote Tool Use, Medical, Law, Finance, and IF.}
\label{tab:merge_schedules}
\small
\begin{tabular}{cccccc}
\toprule
Size & $\gamma$ & $\delta$ & Update 25 & Update 50 & Update 75 \\
\midrule
4B & 0.6 & 0.2 & T,M,L,I & T,M & T \\
4B & 0.6 & 0.3 & T,M,L,I & T & T \\
4B & 0.6 & 0.4 & T,M,L,I & T & $\emptyset$ \\
\midrule
1.7B & 0.4 & 0.1 & T,M,L,I & T,M & T \\
1.7B & 0.4 & 0.2 & T,M,F,I & T & T \\
1.7B & 0.4 & 0.3 & T,M,I & T & T,I \\
1.7B & 0.6 & 0.1 & T,M,L,F,I & T,M,I & T,M,I \\
1.7B & 0.6 & 0.2 & T,M,L,F,I & T,M,I & T \\
1.7B & 0.6 & 0.3 & T,M,L,F,I & T,M,I & T,F,I \\
\bottomrule
\end{tabular}
\end{table}

\subsection{Evaluation Protocol}
\label{app:eval_details}

Evaluation uses temperature 1, top-$p$ 1, and disabled top-$k$. Single-turn tasks use zero presence penalty: MedQA/CaseHOLD/FinQA have 1,273/5,314/1,147 examples and a 26,624-token completion cap; IFBench has 300 prompts and a 32,768-token cap.
We use MedQA, CaseHOLD, FinQA, IFBench, and IFEval for the non-tool domains.\footnote{\url{https://huggingface.co/datasets/GBaker/MedQA-USMLE-4-options}\\
\url{https://huggingface.co/datasets/casehold/casehold}\\
\url{https://github.com/czyssrs/FinQA}\\
\url{https://huggingface.co/datasets/allenai/IFBench_test}\\
\url{https://huggingface.co/datasets/google/IFEval}}
Telecom evaluation uses the official v1.0.1 harness\footnote{\url{https://github.com/sierra-research/tau2-bench}} on 114 base-split tasks, with a 32,768-token context and a 200-step limit. The user simulator is \texttt{gpt-4.1-2025-04-14} at temperature 0; empty responses and execution or context-limit errors count as failures. The split includes 20 task IDs flagged for upstream training overlap, so teacher-data filtering does not exclude upstream exposure.
\looseness=-1

% ==================================================
%                   Next Section
% ==================================================
\clearpage
\section{Additional Experiments}
\label{app:add}

\subsection{Initialization Comparison}
\label{app:initialization_comparison}
\tabautoref{tab:initialization_comparison} gives the five-domain results underlying \figautoref{fig:main}. The $\lambda=1.0$ merge assigns coefficient $0.2$ to each teacher; the $\lambda=2.1$ merge uses $(0.5,0.3,0.2,0.3,0.8)$ in the order Medical, Law, Finance, IF, and Tool Use. SFT warm-up and IM-MOPD follow Appendices~\ref{app:warmup_details} and~\ref{app:merge_details}, respectively. All post-MOPD scores use 100 updates.
\looseness=-1

\begin{table}[ht]
\centering
\caption{Five-domain performance for the initializations in \figautoref{fig:main}. Each entry shows before / after MOPD. IF is evaluated with IFEval prompt-level strict accuracy.}
\label{tab:initialization_comparison}
\small
\begin{tabular*}{\linewidth}{@{\extracolsep{\fill}}lrrrrr@{}}
\toprule
Initialization & MedQA & CaseHOLD & FinQA & IFEval & $\tau^2$ \\
\midrule
No init & 69.5 / 69.1 & 61.8 / 64.7 & 59.1 / 74.3 & 37.5 / 74.9 & 5.3 / 7.9 \\
SFT warm-up & 72.6 / 73.9 & 65.4 / 68.2 & 69.1 / 71.1 & 56.6 / 71.0 & 30.7 / 31.6 \\
Merge ($\lambda=1.0$) & 71.7 / 72.7 & 66.5 / 67.6 & 69.0 / 74.7 & 52.5 / 74.7 & 0.9 / 11.4 \\
Merge ($\lambda=2.1$) & 73.7 / 77.5 & 67.9 / 68.9 & 66.0 / 73.5 & 52.7 / 66.2 & 13.2 / 50.9 \\
IM-MOPD & 71.7 / 76.6 & 66.5 / 70.9 & 69.0 / 74.0 & 52.5 / 75.0 & 0.9 / 69.3 \\
\bottomrule
\end{tabular*}
\end{table}

\subsection{Ablation Details}
\label{app:ablation_results}

\tabautoref{tab:full_ramp4b}--\ref{tab:full_ramp17} report the complete benchmark results underlying the main-text ablations. Both scales are analyzed using the validation recovery rule described in \appautoref{app:merge_details}.

\begin{table}[ht]
\centering
\caption{4B correction-size sweep with $\gamma=0.6$.}
\label{tab:full_ramp4b}
\small
\setlength{\tabcolsep}{3pt}
\begin{tabular}{lrrrrrr}
\toprule
Correction & MedQA & CaseHOLD & FinQA & IFBench & $\tau^2$ & Norm. \\
\midrule
$\delta=0.2$ & 79.1 & 70.3 & 73.9 & 58.7 & 36.8 & 80.3 \\
$\delta=0.3$ & 76.6 & 70.9 & 74.0 & 57.8 & 69.3 & 86.9 \\
$\delta=0.4$ & 78.9 & 71.8 & 73.0 & 54.2 & 59.6 & 86.1 \\
\bottomrule
\end{tabular}
\end{table}

\begin{table}[ht]
\centering
\caption{1.7B recovery-threshold and correction-size sweep.}
\label{tab:full_ramp17}
\small
\setlength{\tabcolsep}{3pt}
\begin{tabular}{lrrrrrr}
\toprule
$(\gamma,\delta)$ & MedQA & CaseHOLD & FinQA & IFBench & $\tau^2$ & Norm. \\
\midrule
$(0.4,0.1)$ & 53.7 & 62.9 & 59.0 & 34.8 & 14.0 & 62.2 \\
$(0.4,0.2)$ & 52.2 & 59.5 & 59.3 & 34.0 & 14.0 & 57.2 \\
$(0.4,0.3)$ & 53.7 & 59.4 & 55.9 & 42.9 & 28.1 & 76.0 \\
$(0.6,0.1)$ & 54.6 & 62.1 & 58.6 & 35.2 & \,\,\,8.8 & 57.7 \\
$(0.6,0.2)$ & 55.0 & 63.9 & 58.2 & 38.4 & 12.3 & 65.0 \\
$(0.6,0.3)$ & 56.9 & 64.7 & 50.5 & 41.8 & 19.3 & 69.8 \\
\bottomrule
\end{tabular}
\end{table}

% \clearpage
\subsection{Student-prefix Teacher-continuation Diagnostics}
\label{app:prefix_teacher}

We vary the student-generated fraction $\alpha\in\{0.1,0.3,0.5,0.7\}$; the 100\% endpoints are evaluated using the student alone.
\looseness=-1

\paragraph{Medical.}
For each recorded MedQA response of length $L$, the teacher continues after the first $\lfloor\alpha L\rfloor$ student-generated tokens. The prefix length is therefore determined separately for each sample.
\looseness=-1

\vspace{-10pt}
\paragraph{Tool Use.}
In $\tau^2$-Telecom, we instead use a per-turn budget $N=\operatorname{round}(\alpha\bar L)$, where $\bar L$ is the student's mean tokens per assistant turn in a standalone evaluation. The teacher continues when the student reaches this budget. We use the mean because changing one turn in a multi-turn simulation changes subsequent interactions, so later turns cannot be matched to their original sample-specific lengths. Because the budget uses the mean turn length, some shorter turns may be generated entirely by the student.
\looseness=-1

% \clearpage
\subsection{Two-domain Initialization and Merge Timing}
\label{app:two_domain}

\tabautoref{tab:two_domain} summarizes the MedQA and IFEval results. IFEval reports prompt-level strict accuracy over 541 prompts. Norm. averages normalized recovery over these two benchmarks.
\looseness=-1

\begin{table}[ht]
\centering
\caption{Two-domain performance before and after MOPD. Simplex coefficients are ordered as Medical and IF. The last row uses uniform initialization and mid-training merging.}
\label{tab:two_domain}
\small
\setlength{\tabcolsep}{4pt}
\begin{tabular}{lrrrrrr}
\toprule
& \multicolumn{3}{c}{Initialization} & \multicolumn{3}{c}{After MOPD} \\
\cmidrule(lr){2-4}\cmidrule(lr){5-7}
Initialization / method & MedQA & IFEval & Norm. & MedQA & IFEval & Norm. \\
\midrule
$(0.2,0.8)$ & 73.4 & 79.7 & 76.7 & 76.0 & 78.4 & 86.4 \\
$(0.4,0.6)$ & 75.6 & 79.7 & 86.7 & 76.7 & 79.5 & 91.3 \\
$(0.6,0.4)$ & 77.4 & 64.5 & 75.4 & 79.4 & 79.1 & 102.5 \\
$(0.8,0.2)$ & 80.0 & 49.9 & 68.9 & 80.3 & 78.4 & 105.4 \\
IM-MOPD & 78.3 & 68.8 & 84.8 & 80.0 & 79.3 & 105.2 \\
\bottomrule
\end{tabular}
\end{table}

}

\end{document}